\pdfoutput=1
\documentclass[10pt,twocolumn,letterpaper]{article}

\usepackage[pagenumbers]{cvpr} 

\usepackage{graphicx}
\usepackage{amsmath}
\usepackage{amssymb}
\usepackage{booktabs}

\usepackage[pagebackref,breaklinks,colorlinks]{hyperref}

\usepackage[capitalize]{cleveref}
\crefname{section}{Sec.}{Secs.}
\Crefname{section}{Section}{Sections}
\Crefname{table}{Table}{Tables}
\crefname{table}{Tab.}{Tabs.}

\def\cvprPaperID{*****} 
\def\confName{CVPR}
\def\confYear{2023}

\begin{document}

\title{A Dual Branch Network for Emotional Reaction Intensity Estimation}

\author{
Jun Yu, Jichao Zhu, Wangyuan Zhu, Zhongpeng Cai, Guochen Xie, Renda Li, Gongpeng Zhao\\
University of Science and Technology of China\\
{\tt\small harryjun@ustc.edu.cn},\\ 
{\tt\small \{jichaozhu, zhuwangyuan, zpcai, xiegc, rdli, zgp0531\}@mail.ustc.edu.cn}
}
\maketitle

\begin{abstract}
Emotional Reaction Intensity(ERI) estimation is an important task in multimodal scenarios, and has fundamental applications in medicine, safe driving and other fields. In this paper, we propose a solution to the ERI challenge of the fifth  Affective Behavior Analysis in-the-wild(ABAW), a dual-branch based multi-output regression model. The spatial attention mechanism is used to better extract visual features, and the Mel-Frequency Cepstral Coefficients technology extracts acoustic features, and a method named modality dropout is added to fusion multimodal features. Our method achieves excellent results on the official validation set.
\end{abstract}

\section{Introduction}
\label{sec:intro}

With advances in artificial intelligence and deep learning, researchers are increasingly interested in computational methods for human emotional reactions\cite{Christ2022TheM2}. It can help doctors diagnose whether patients have anxiety, depression, etc. by calculating the intensity of emotional reactions. In addition, it can also be used in scenarios such as education\cite{Wilhelm2019TowardsFE}, entertainment\cite{Akbar2019EnhancingGE}, and driver safety detection\cite{Jeong2018DriversFE}.

For static facial expression recognition and dynamic facial expression recognition tasks, common emotion description methods include action units (AU), arousal and valence, etc, which have been used in ABAW's challenge \cite{kollias2022eccv, kollias2022cvpr, kollias2021distribution, kollias2021analysing, kollias2021affect, kollias2020analysing, kollias2019expression, kollias2019face, kollias2019deep, zafeiriou2017aff}.

For traditional methods based on Support Vector Machine-Hidden Markov Model(HMM)\cite{Krishna2013EmotionRU}. With the development of deep learning, CNN, 3DCNN, and RNN methods have been applied to visual tasks. In recent years, with the excellent performance of Transformer\cite{Vaswani2017AttentionIA} in natural language processing, ViT\cite{dosovitskiy2020image} has been successfully applied in computer vision, and has produced many excellent pre-training models. However, these works mainly classify emotional samples into specific, which is a typical single classification task. Furthermore, datasets collected in laboratory share similar fixed patterns, with emotional expressions having similar intensities. By collecting data on the web and creating datasets to make it more wild. However, in these tasks, static or dynamic expressions are recognized as limited class, while the connection between emotions is ignored \cite{Wang2022EmotionalRA}, which is not enough to finely reflect the emotional state.

In order to promote the development of emotional reaction intensity(ERI), ABAW2023 holds this competition, our goal is to design a model to predict the reaction intensity of different emotions, including: Adoration, Amusement, Anxiety, Disgust, Empathic-Pain, Fear, Surprise, so it's a multi-output regression task.
Therefore, in this paper, we propose a video feature extraction model based on CNN and spatial attention. On the basis of fusing local inter-frame information through temporal convolutional network, we use the temporal transformer to obtain the global temporal relationship, and use the MFCC to generate audio branch's feature. Use a timing modeling method similar to video branch to obtain the global timing relationship. Finally, the features of the video and audio are fused into the prediction head to estimate the vector of emotional reaction intensity.

In summary, our contributions can be summarized as follows:
\begin{itemize}
    \item We propose a dual-branch model for the ABAW ERI Estimation track. It consists of  Spatial Encoder with CNN, MFCC and Temporal Encoder.
    \item We introduce a mechanism based on modality dropout to fuse visual and aduio features.
\end{itemize}

\section{Related Work}
\label{sec:related}
In \cite{Vaiani2022ViPERVP} they propose a method called ViPER. Based on the pre-trained Vision Transformer\cite{Dosovitskiy2020AnII}, a modality-independent fusion framework is designed to predict people's emotional state, whose input data can be a combination of audio and video frames and text. For dynamic facial 

\begin{figure*}[ht]
	\centering
    \includegraphics[width=17cm]{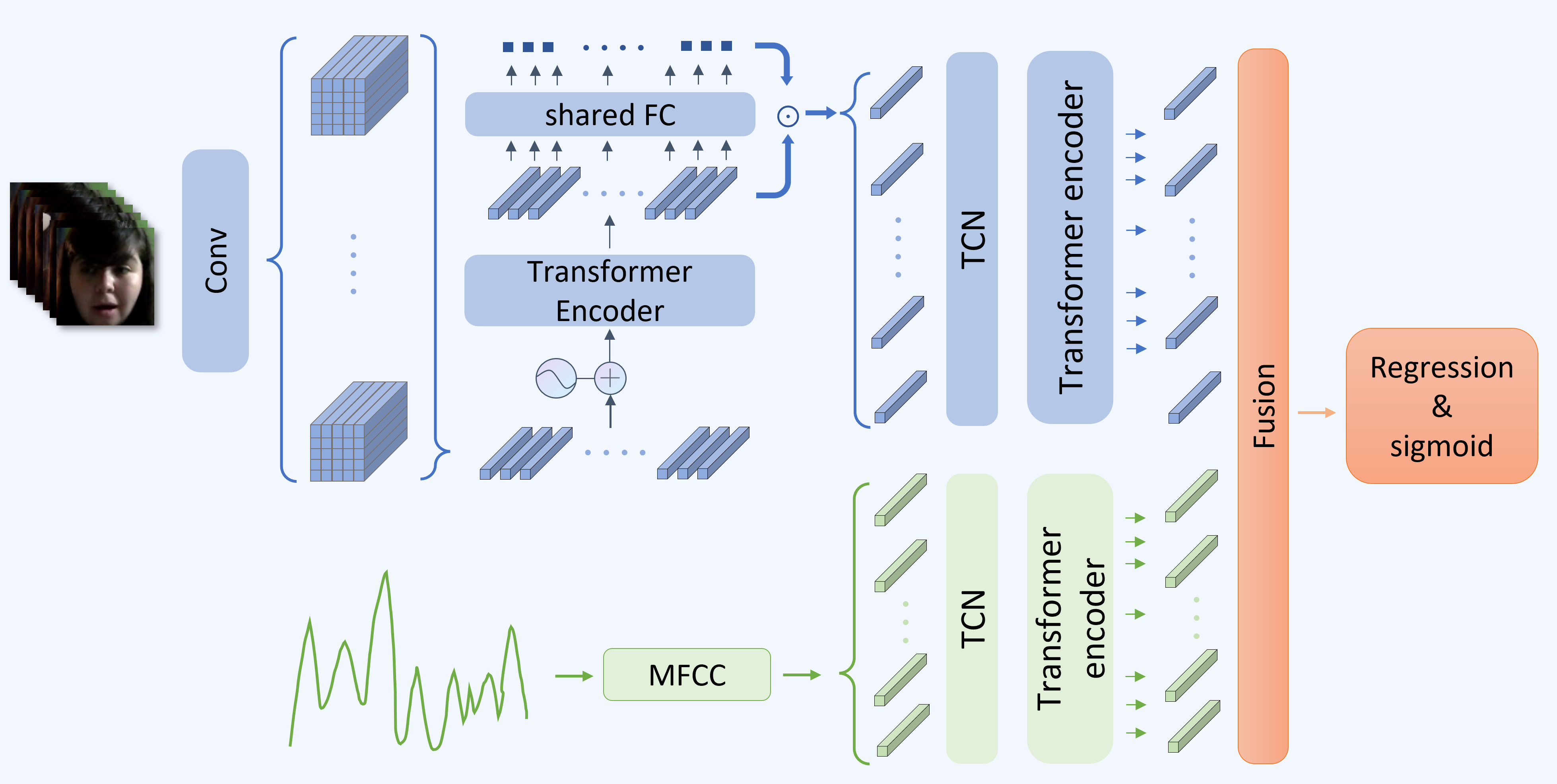}
    \caption{proposed architecture}
    \label{fig:model}
\end{figure*}
\noindent

\noindent
expression recognition, unlike static facial expression recognition samples which tend to exhibit consistent high-intensity expressions, \cite{Li2022IntensityAwareLF} designs a global convolution-attention block (GCA) module to weight the feature map channels so as not to Correlated features can be suppressed to avoid equal treatment of frames with different expression strengths in a video sequence. Meanwhile, an intensity-aware loss-guided network is designed to distinguish emotion samples with relatively low expression intensity.
Feature extraction from video streams affects model performance. \cite{Wang2022EmotionalRA, KR2021SubwordLL} both proposed models based on spatial attention mechanism and aggregating different frames through convolutional or linear layers. Among them, \cite{Wang2022EmotionalRA} is pre-trained on the DFEW dataset and trained on ERI, which improves the effect of processing dynamic emotions. 
For audio-based ERI estimation, \cite{Li2022HybridMF, Wang2022EmotionalRA} use  extended Geneva Minimalistic Acoustic Parameter Set(eGeMAPS) and DeepSpectrum of DenseNet121 pretrained on ImageNet as features, but the performance are lower than video-based methods. \cite{Li2022HybridMF} also uses ResNet18\cite{He2015DeepRL} as the backbone to extract audio features, achieving the best results based on audio methods.

\section{Methodology}
In this section, we describe our method in detail, the architecture of the model is shown in \cref{fig:model}

\label{sec:method}
\subsection{Pre-processing}
It is necessary to preprocess the video streams in the dataset, since the focus of ERI is the facial region, which should be avoided from being disturbed by other factors. The videos are first split into images, and we use the dlib\cite{King2009DlibmlAM} toolkit to detect 68 facial landmarks in each frame to crop the face, and resize to $112\times112$ as the input size. Besides, interpolation with a window width of 12 frames and frame smoothing are used to handle frames that cannot be detected by dlib. For face samples cannot be detected in the entire video frame, we also use a more robust MTCNN\cite{Zhang2016JointFD} method. Due to the subject's webcam has different fps, the frame number of the video ranges from $50$ to $1561$, while the duration of the video ranges from $9.9$ seconds to $15$ seconds. Therefore, in order to facilitate processing and save GPU memory, we uniformly extract $32$ frames from each sample as input.

\subsection{Visual backbone}
\textbf{CNN.} 
The input $x$ for the visual branch is faces clipped from the original video stream with a linear sampling of $T=32$ frames, for our model $x \in \mathbb{R}^{T\times H \times W \times 3}$. 
The first foure convolutional layers of ResNet18\cite{He2015DeepRL} are applied on cropped images to extract low-level features $f\in \mathbb{R}^{T\times c \times h \times w}$.

\textbf{Spatial-Encoder.} In order to input the CNN feature map $f$ of every input frame $t\in \{1, 2, \cdots, T\}$ into a shared spatial-encoder, we first flatten each frame's feature into a two-dimensional shape as $f_{t}\in \mathbb{R}^{(hw)\times d_{model}}$, and add the spatial positional embedding to $f_{t}$. Therefore, the encoder's input can be defined as: 
\begin{equation}
f_{t, i} =   f_{t,i} + p_{i}
\end{equation}
where $p_{i} (i\in \{1, 2, \cdots, hw\})$ is a learnable location parameter, and an enhanced feature map $z_{t}$ is obtained after computing by spatial encoder as:
\begin{equation}
    z_{t} = \text{encoder}(f_{t})
\end{equation}
A full-connection layer with softmax is added on $z_{t}$ to aggregate the information of each position and generate a position weight $a_{t}$. Weighted by $a_{t}$, from the enhanced feature map $z_{t}$ , we obtain aggregation feature $g_{t}$:
\begin{align}
a_{t, i} &= \text{FC}(z_{t, i}), i\in\{1, 2, \cdots, hw\} \\
a_{t} &= \text{softmax}(a_{t, 1}, \cdots, a_{t, hw}) \\
g_{t} &= \sum_{i=1}^{hw}(a_{t}z_{t})
\end{align}
where FC denotes full-connection layer, and $g_{t}$ is corresponded to a frame. By stacking $g_{t}$ in the time dimension, we obtain the feature sequence $g=(g_{1}, \cdots, g_{T})$ of a video only containing the position representation.

\subsection{Acoustic feature}
Mel Frequency Cepstral Coefficient(MFCC) is an audio feature widely used in speech recognition and emotion recognition, which is very close to the human hearing system. We used the Python toolkit Librosa to extract 128-dimensional features, and then combined the adjacent 8-frame features to obtain 1024-dimensional feature vector, which is fed into temporal encoder described in the next.

\subsection{Temporal Encoder}
In this section, the module proposed by \cite{Chen2021TransformerEW} is adopted. The extracted feature vectors from visual or audio backbone are fed into the temporal convolutional network(TCN) based on 1-dimensional causal convolutional with dilation to aggregate local temporal context . Besides, zero padding is also used to ensure that the output of the feature through the convolutional layer has the same length as the input. Causal convolution can ensure that the current time $t$ can only see the information at the previous time, so as to avoid information leakage. Denote $ker$ as the convolution kernel and dilation rate is $d$, and the input of time dimension $x$, then the $p$-th element of output feature $y$ can be calculated by:
\begin{equation}
   y_{p}=\sum_{di+t=p}ker(i)\cdot x_{t}
\end{equation}
by using different $d$, the receptive field of the convolution can be changed dynamically. Finally, add temporal position information to the output of TCN as input of transformer encoder to capture global information.

\subsection{Multimodal Fusion and Regression}
To efficiently integrate the good features learned in video and audio models, and to avoid the model's excessive dependence on a certain modality during learning, we use a fusion strategy called modality dropout, which is applied at the modal level. With a probability $p_m$, both audio and video feature are used as input, when only one is used, the video feature is selected with a probability of $p_v$. Given the audio feature $f^{a}$ and video feature $f^{v}$, the mltimodal feature $f^{m}$with modality dropout is:
\begin{equation}
    f^{av} = 
    \begin{cases}
    \text{concat}(f^{a}, f^{v}) & \text{with } p_m\\
    \text{concat}(0, f^{v}) & \text{with } (1-p_m)p_{v}\\
    \text{concat}(f^{a}, 0) & \text{with } (1-p_m)(1-p_{v})\\
    \end{cases}
\end{equation}
\begin{equation}
    f^{m} = W_{m}f^{av} + b_m
\end{equation}
where $W_m$ and $b_m$ are learnable parameters, concat denotes channel-wise concatenation, and a layer normalization \cite{Ba2016LayerN} is added after concatenation.
Finally, similar aggregation method in spatial encoder is used to estimate the reaction intensity. The difference is that sigmoid is added as the activation function to normalize the value to $(0, 1)$.

\subsection{Optimisation objective}
In this work, we use mean square error(MSE) loss for our training process. Let $y=[y_{1}, \cdots, y_{7}]$ and $\hat{y}=[\hat{y}_{1}, \cdots, \hat{y}_{7}]$ be the true emotional reaction intensity and the prediction, respectively, then the loss $\mathcal{L}$ can be defined as:
\begin{equation}
\mathcal{L} = \text{MSE}(y, \hat{y}) = \mathbb{E}[\sum_{i,j}(y_{ij}-\hat{y}_{ij})^{2}]
\end{equation}
where $i$ denotes the emotion, $j$ denotes the batch size.


\begin{figure*}[htbp]
  \begin{minipage}{0.5\textwidth}
    \centering
    \includegraphics[width=\linewidth]{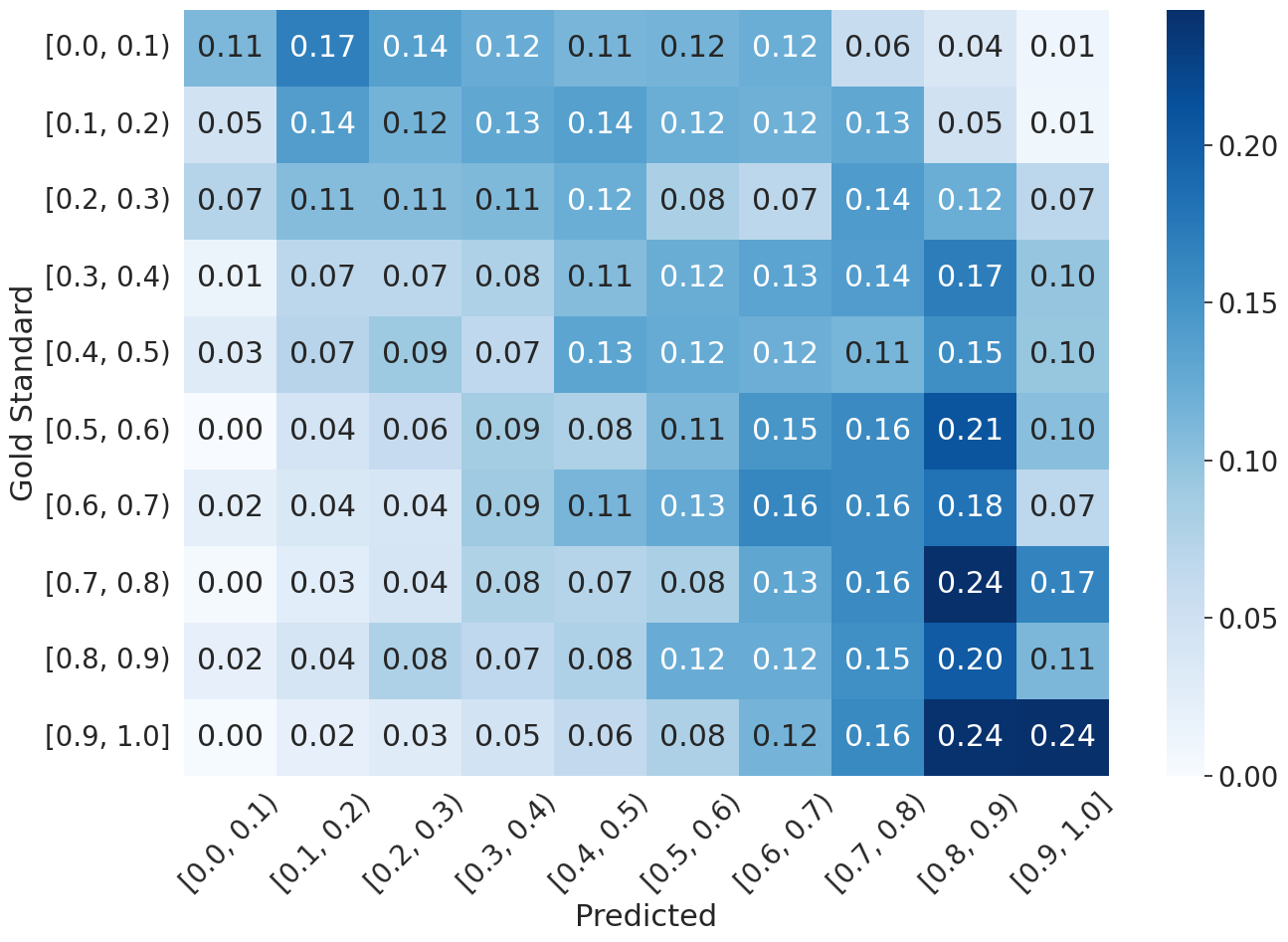}
    \caption{Amusement prediction analysis}
    \label{fig:amusement}
  \end{minipage}%
  \begin{minipage}{0.5\textwidth}
    \centering
    \includegraphics[width=\linewidth]{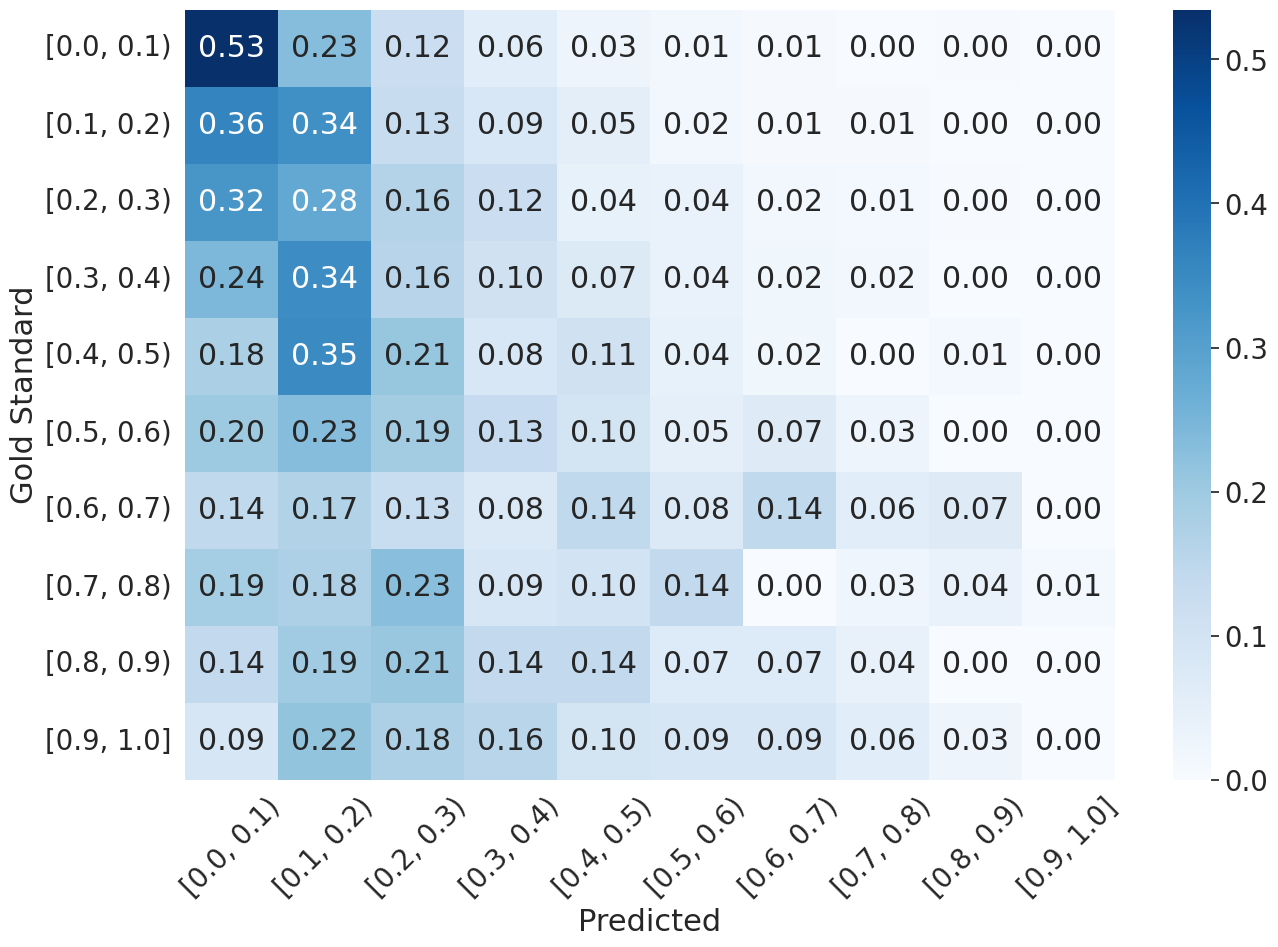}
    \caption{Amusement prediction analysis}
    \label{fig:pain}
  \end{minipage}
\end{figure*}

\section{Experiment}
\label{sec:exp}
\subsection{Dataset}
Hume-Reaction dataset is used for the ERI Estimation Challenge in 5th ABAW. It is a reaction of subjects from two cultures, South Africa and the United States, to emotional video stimuli. It consists of both audio and video parts and is recorded over approximately $75$ hours. Corresponding label vector is self-annotated by the subjects and normalized to $[0, 1]$ by its maximum intensity, and $7$ elements represent adoration, amusement, anxiety, disgust, empathic-pain, fear, and surprise, respectively.

\subsection{Implement Details}

\textbf{Evaluation metric} Average pearson’s correlations coefficient ($\rho$) is the metric used in intensity estimation, which is a measure of linear correlation between predicted emotional reaction intensity and target, then the metric can be defined as follows:
\begin{equation}
\bar{\rho} = \sum_{i=1}^{7} \frac{\rho_{i}}{7}
\end{equation}
where $\rho_{i}(i \in \{1,2\cdots,7\})$ for $7$ emotions, respectively, and is defined as:
\begin{equation}
\rho_{i} = \frac{\text{cov}(y_{i}, \hat{y_{i}})}{\sqrt{\text{var}(y_{i})\text{var}(\hat{y_{i}})}}
\end{equation}
where $\text{cov}(y_{i}, \hat{y_{i}})$ is the covariance between the predicted value and the target, $\text{var}(y_{i})$ and $\text{var}(\hat{y_{i}})$ are variance respectively.

\textbf{Training settings} The training process is optimized by Adam \cite{Kingma2014AdamAM} optimizer. All the experiments are implemented on NVIDIA RTX 3090 with PyTorch, with initial learning rate of $e^{-4}$, batch size of $64$. And when metric on the validation set don't improve for $10$ epochs, the learning rate will halved. For visual branch, we use ResNet18 weight from \cite{Wen2021DistractYA} trained on AffectNet\cite{Mollahosseini2017AffectNetAD} as initialization parameter, it can capture effective features in static face representation recognition task. We freeze the unimodal's model parameters with the highest $\rho$, extract audio and video features and feed them to the fusion module. The dimension of encoder in visual branch is $256$, equals to low level feature's channel, and the number of encoder blocks is $4$ and number of multi-head is $4$. In temporal encoder, the kernel size of 1-dimension convolution is $3$, and convolution layer is $5$, the dimension of feature in attention is $128$. For fusion module, modality dropout and video dropout are set to $0.1$ and $0.5$, respectively.

\subsection{Results}
In our unimodal experiment, both the branch based on video and audio, the performance of the model are greatly improved as shown in \cref{tab:dev set}, and the correlation coefficients are increased to $0.2972$, $0.3500$, which have a significant improvement compared with baseline.
\begin{table}
  \centering
  \begin{tabular}{l|c|c|c}
    \toprule
    Method & Audio & Video & $\bar{\rho}$\\
    \midrule
    Baseline(eGeMAPS)\cite{Kollias2023ABAWVE}      & \checkmark & -             & 0.0583\\
    Baseline(DeepSpectrum)\cite{Kollias2023ABAWVE} & \checkmark & -             & 0.1087\\
    ours                                            & \checkmark & -             & 0.2972\\
    Baseline(FAU)\cite{Kollias2023ABAWVE}          &   -        & \checkmark    & 0.2840\\
    Baseline(VGGFACE2)\cite{Kollias2023ABAWVE}     &   -        & \checkmark    & 0.2488\\
    ours                                            &   -        & \checkmark    & 0.3500\\
    Baseline\cite{Kollias2023ABAWVE}               & \checkmark & \checkmark    & 0.2382\\
    ViPER\cite{Vaiani2022ViPERVP}                  & \checkmark & \checkmark    & 0.3025\\
    ours                                             & \checkmark & \checkmark    & 0.4429\\
    \bottomrule
  \end{tabular}
  \caption{Results on validation set}
  \label{tab:dev set}
\end{table}

\textbf{Ablations}
We conducted ablation studies on the Hume-Reaction dataset to better understand our proposed model, and the results are shown in \cref{tab:fusion module}. If $p_{m}$ is set to $1.0$, i.e., the two features from temporal encoder are directly concatenated together in last dimension, $\rho$ is $0.4426$, is significantly boosted compared with unimodal, $p_m=0.9$, $\rho$ has a slight improvement.

\begin{figure}[ht]
	\centering
    \includegraphics[width=0.45\textwidth]{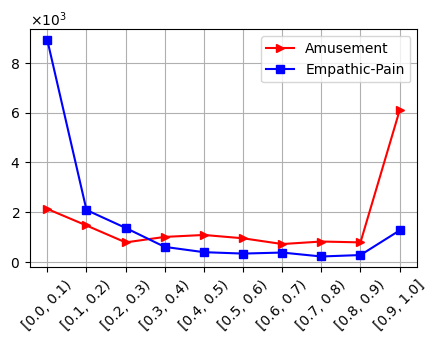}
    \caption{Amusement and empathic-pain sample statistics with quantized intensity on train set}
    \label{fig:quantized}
\end{figure}

\textbf{Qualitative analysis} On the validation set, we calculated the emotional response intensity of different samples, and the average value was $0.44$, which indicated that our estimated value could well reflect the emotions of the sample subjects.
For the convenience of analysis, we selected amusement and empathic-pain according to the average response intensity of each emotion in the training set, divided them into 10 levels on the basis of intensity, and counted the corresponding number of samples. The results are shown in \cref{fig:quantized}. The former has more high-intensity samples, while the latter has a majority of low-intensity samples. 

\cref{fig:amusement} and \cref{fig:pain} show the confusion matrix between the estimated value and the actual intensity of the amusement class and empathic-pain class, respectively. For the amusement class, these values are roughly scattered around the diagonal, and perform well for high intensities with $0.28$, indicating that the model can calculate the intensity of emotional responses well, while the empathic-pain class is mainly distributed in $[0, 0.3]$. This difference may be caused by the label distribution of the train set, as shown in \cref{fig:quantized}, the label intensity of the latter is mainly concentrated in $[0, 0.2]$, or consistent with the classification results in \cite{Jiang2020DFEWAL, Li2022IntensityAwareLF} , the amusement(happy) class is easier to be learned by the model.


\section{Conclusion}
\label{sec:conc}

In this paper, we propose a multimodal based method to improve the performance of emotional response intensity estimation, extracting the global feature information of the face based on spatial attention in the visual branch, and generating features based on MFCC in the acoustic branch. Our method outperforms prior work on the Hume-Reaction dataset.

\begin{table}
  \centering
  \begin{tabular}{l|c|c|c}
    \toprule
     Method & $p_{m}$ & $p_{v}$ & $\bar{\rho}$  \\
    \midrule
    ours & 1.0 & -   & 0.4426 \\
    ours & 0.9 & 0.5 & 0.4429 \\
    ours & 0.8 & 0.5 & 0.4424 \\
    \bottomrule
  \end{tabular}
  \caption{Result with modality dropout on validation set}
  \label{tab:fusion module}
\end{table}


\section*{Acknowledge}
Natural Science Foundation of China (62276242), CAAI-Huawei MindSpore Open Fund (CAAIXSJLJJ-2021-016B, CAAIXSJLJJ-2022-001A), Anhui Province Key Research and Development Program (202104a05020007), USTC-IAT Application Sci. \& Tech. Achievement Cultivation Program (JL06521001Y), Sci.\&Tech. Innovation Special Zone (20-163-14-LZ-001-004-01).

{\small
\bibliographystyle{ieee_fullname}
\bibliography{egbib}
}

\end{document}